\documentclass[letterpaper, 10 pt, conference]{ieeeconf}
\IEEEoverridecommandlockouts
\usepackage{cite}
\usepackage{amsmath,amssymb,amsfonts}
\usepackage{algorithmic}
\usepackage{graphicx}
\usepackage{textcomp}
\usepackage{xcolor}
\usepackage{subfig}
\usepackage{bm}
\usepackage{pifont}
\usepackage{booktabs}
\usepackage{multirow}
\allowdisplaybreaks[4]

\title{\LARGE \bf
Night-Rider: Nocturnal Vision-aided Localization in Streetlight Maps Using Invariant Extended Kalman Filtering
}

\author{Tianxiao Gao$^{1}$, Mingle Zhao$^{1}$, Chengzhong Xu$^{1}$, and Hui Kong$^{1*}$ 
\thanks{*Corresponding author.}
\thanks{$^{1}$Tianxiao Gao, Mingle Zhao, Chengzhong Xu, and Hui Kong are with the State Key Laboratory of Internet of Things for Smart City (SKL-IOTSC), Faculty of Science and Technology, University of Macau, Macao, China. ({\tt\small \{ga0.tianxiao, zhao.mingle\}@connect.umac.mo, \{czxu, huikong\}@um.edu.mo})}}

\begin{document}


\maketitle
\thispagestyle{empty}
\pagestyle{empty}

\begin{abstract}
Vision-aided localization for low-cost mobile robots in diverse environments has attracted widespread attention recently. Although many current systems are applicable in daytime environments, nocturnal visual localization is still an open problem owing to the lack of stable visual information. An insight from most nocturnal scenes is that the static and bright streetlights are reliable visual information for localization. Hence we propose a nocturnal vision-aided localization system in streetlight maps with a novel data association and matching scheme using object detection methods. We leverage the Invariant Extended Kalman Filter (InEKF) to fuse IMU, odometer, and camera measurements for consistent state estimation at night. Furthermore, a tracking recovery module is also designed for tracking failures. Experimental results indicate that our proposed system achieves accurate and robust localization with less than $0.2\%$ relative error of trajectory length in four nocturnal environments.
\end{abstract}


\section{Introduction}
With the increasingly versatile applications of mobile robots and autonomous vehicles, accurate and robust localization has been in great demand. Since cameras are cost-effective and informative, extensive visual SLAM methods\cite{mur2017orb, engel2017dso, qin2018vins, campos2021orb} have been developed for diverse mobile robots. Nevertheless, dynamic objects make the visual localization without prior maps prone to failures and the online-built local map increases computation costs. Therefore, many visual localization methods\cite{zhang2023cross, wang2021visual, liao2020coarse, guo2021coarse, yu2020monocular, caselitz2016monocular, ma2019exploiting, zhang2019vision_aided} using pre-built maps are proposed recently. These methods can be categorized to feature-based and semantics-based. Feature-based methods \cite{yu2020monocular, caselitz2016monocular, zhang2023cross, zhang2019vision_aided} extract the low-level features and match them with map points for pose estimation. Semantics-based visual localization methods \cite{wang2021visual, liao2020coarse, guo2021coarse, ma2019exploiting} utilize the segmented poles, lanes, signboards, and curbs to estimate the 2D-3D correspondences between images and the map. Although they achieve accurate localization, the vast majority of them only consider localization in the daytime or bright scenarios. Currently, the visual localization in nighttime scenes is still an open problem.

\begin{figure}[t]
    \centerline{\includegraphics[scale=0.40]{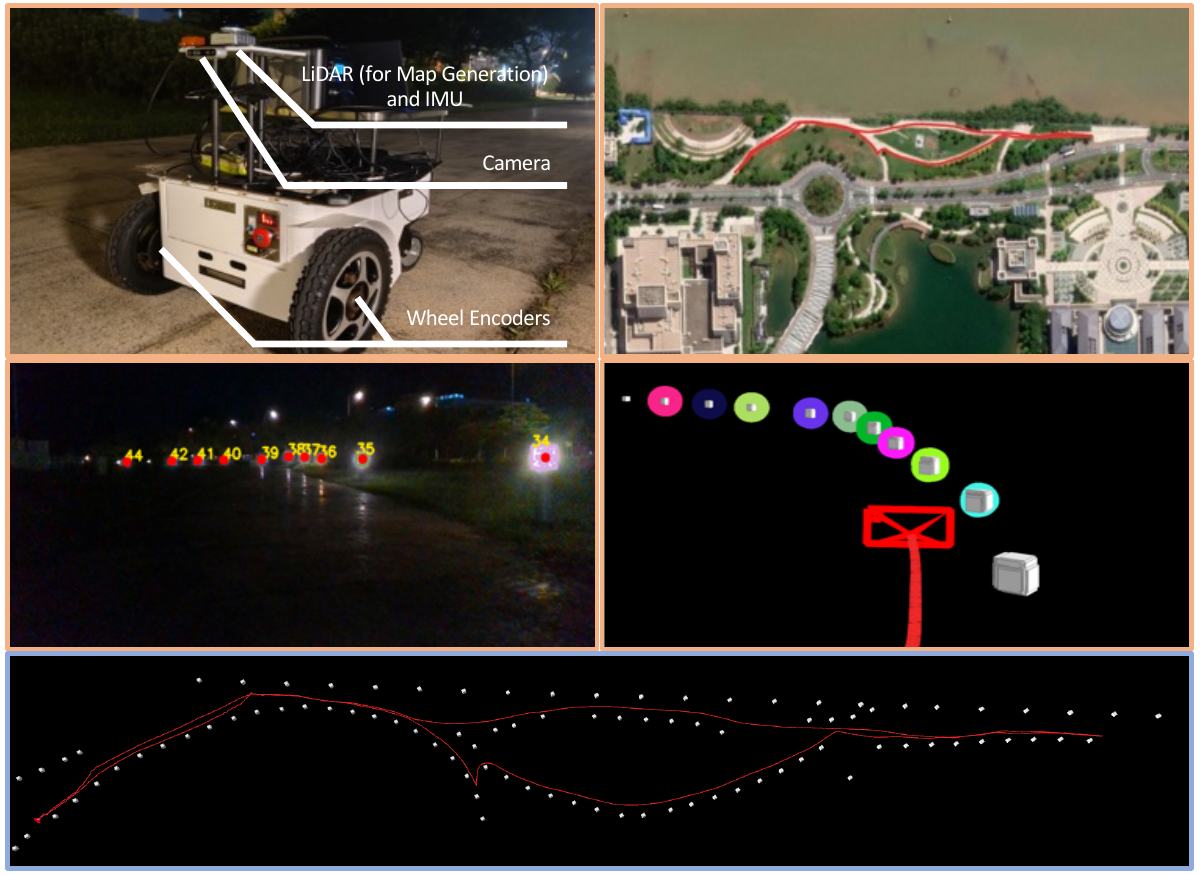}}
    \caption{The proposed nocturnal localization system in the nighttime campus streetlight map. Top-left image shows the robot configuration. Top-right image is the trajectory (red) overlaid with the daytime satellite map for visualization. The middle images display the matches of 2D streetlight detections and 3D streetlight clusters. The bottom image depicts the estimated trajectory in the streetlight map.}
    \label{preview}
\end{figure}

One problem of visual localization in nocturnal environments is the drastically changed luminance. 
The background is illuminated by bright light sources and becomes completely dark in the absence of light. 
Such a problem makes the current visual localization methods not applicable to nocturnal scenes. For feature-based visual localization methods, low-level feature extraction is not effective at night as it is sensitive to light conditions. The unstable luminance at night makes the semantics-based methods become invalid as well\cite{gao2022cross}. Fortunately, there are abundant static artificial streetlights that can provide a source of rare perceptive information in most nocturnal scenarios (e.g. nighttime campuses or parks). Additionally, the light saturation makes streetlights easy to detect from images. Therefore, the purpose of this paper is to localize mobile robots in nocturnal environments by leveraging streetlights to establish connections between the camera measurements and the 3D environment.

\begin{figure*}[htbp]
	\centerline{\includegraphics[scale=0.45]{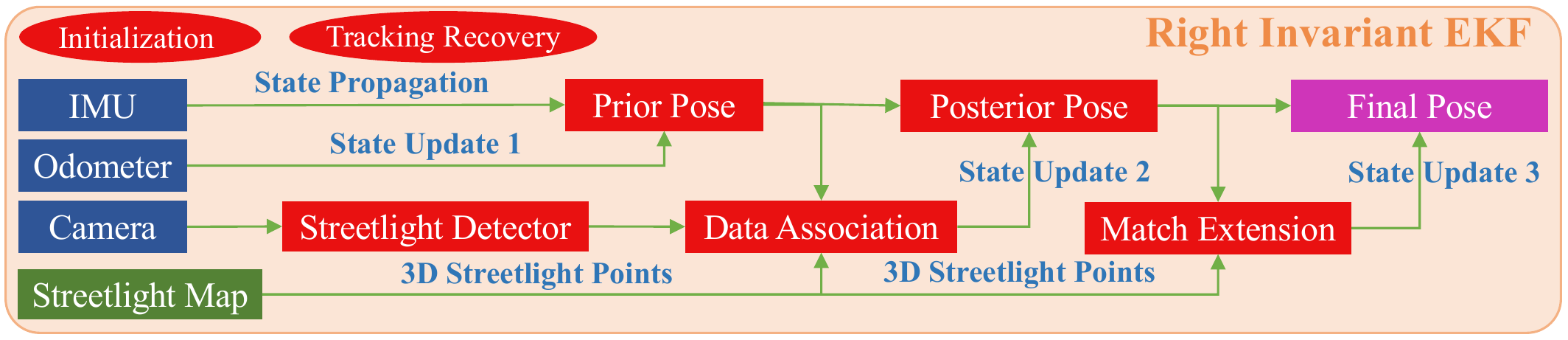}}
	\caption{System overview of the nocturnal vision-aided localization system. The InEKF is utilized for state estimation.}
	\label{system_overview}
\end{figure*}

In this paper, we propose an accurate and robust nocturnal vision-aided localization system in streetlight maps based on an Inertial Measurement Unit (IMU), an odometer, and a monocular camera. 
Note that NightNav\cite{nelson2015dusk} has a similar idea of using streetlights for nighttime visual localization. However, NightNav introduces constraints for pose optimization by assuming that the area of streetlights is proportional to the inverse of the square of the distance between the camera and streetlights, which is not always true as the angle of view, occlusion, and overlap of lights can significantly interrupt the relation. In contrast, in our system, there is no assumption about nocturnal scenes. Besides, the pre-stored images of all streetlights in NightNav drastically increase the storage cost while our system only requires lightweight maps with sparse streetlight points. 
Fig. \ref{preview} shows the robot configuration and an example of the estimated trajectory in the streetlight map. The main contributions of the paper are: 
\begin{itemize}
   \item We propose a real-time nocturnal vision-aided localization system, Night-Rider. The system can achieve accurate and robust localization by leveraging limited perceptive information from streetlights at night.
    
    \item We propose a novel matching scheme for data association in streetlight maps which enables 6-DoF localization with extremely low drift in nighttime scenarios. 

    \item To mitigate potential errors in nocturnal scenes, the Invariant Extended Kalman Filter (InEKF) is introduced for consistent and stable state estimation. 
    
    \item We perform extensive experiments on a wheeled robot in real nighttime scenes to validate the effectiveness, accuracy, and robustness of the proposed system. 
\end{itemize}

\section{System Overview}
Fig. \ref{system_overview} shows the framework of the proposed nocturnal localization system. The streetlight map is constructed using only the 3D streetlight points. 
We introduce the right InEKF due to its accuracy and consistency in the system for fusing monocular camera, IMU, and odometer measurements. Specifically, the robot state is propagated using IMU measurements, and the odometer measurements are utilized for the first state update. Next, the updated states are fed into the proposed data association module to establish correspondences between 3D streetlight clusters and 2D detections. The matches will be used for the second state update. More 2D-3D streetlight matches are then obtained through the binary segmentation in the match extension module and are applied to the third state update. Additionally, a tracking recovery module is proposed to resolve localization failures resulting from insufficient matches.

As for the notations and frames, $(\cdot)^{w}$, $(\cdot)^{c}$, $(\cdot)^{o}$ and $(\cdot)^{b}$ denote the world, camera, odometer and body frames. The IMU frame coincides with the body frame. The first IMU frame is regarded as the world frame and the gravity vector $\mathbf{g}^{w}$ is constant in the world frame. 
$\mathbf{R}^{w}_{b} \in SO(3)$ and $\mathbf{p}^{w}_{b} \in \mathbb{R}^{3}$ are rotation and translation from the world frame to body frame. All velocities in this paper are with respect to the world frame. For example, $^{w}\mathbf{v}_{b} \in \mathbb{R}^{3}$ is the velocity of body frame with respect to and represented in the world frame. $\Tilde{(\cdot)}$ and $\hat{(\cdot)}$ represent the measurement and the estimate, respectively. 
\section{State Estimation and Localization}
In this section, we first briefly introduce the generation of streetlight map. Then we illustrate the state representation of the system followed by the InEKF for state estimation fusing IMU, odometer, and camera measurements. The proposed data association and match extension methods enable the system to accurately localize 6-DoF poses at night. Finally, we explicate the tracking recovery module for tracking failure cases. 

\subsection{Streetlight Map Generation}\label{mapping}
We employ an image-based object detection method\cite{chen2021yolov5} to filter luminous streetlight points from LiDAR maps or measurements for generating the pre-built streetlight map. For each LiDAR scan, points whose projections fall into the detected streetlight boxes are labeled as “streetlight”. The original LiDAR map is generated by the modified FAST-LIO2\footnote{https://github.com/hku-mars/FAST\_LIO} \cite{xu2022fast} augmented with a fast and accurate loop closure module \cite{lidar_iris} and only the map points with “streetlight” labels are retained. The streetlight point map is further filtered and clustered by DBSCAN\cite{schubert2017dbscan} for grouping 3D points into different streetlight instances. Finally the neat 3D streetlight map can be generated as shown in the bottom of Fig. \ref{preview}.

\subsection{State Representation}
The robot state includes the orientation $\mathbf{R}^{w}_{b} \in SO(3)$, position $\mathbf{p}^{w}_{b}\in \mathbb{R}^{3}$, velocity $^{w}\mathbf{v}_{b} \in \mathbb{R}^{3}$, as well as the IMU accelerometer bias $\mathbf{b}_{a} \in \mathbb{R}^{3}$ and gyroscope bias $\mathbf{b}_{\omega} \in \mathbb{R}^{3}$. 

The orientation $\mathbf{R}^{w}_{b}$, position $\mathbf{p}^{w}_{b}$ and velocity ${^{w}}\mathbf{v}_{b}$ evolve on the matrix Lie group $SE_{2}(3)$ \cite{hartley2020contact}. The element $\mathbf{X}_{t} \in SE_{2}(3)$ of time $t$ is represented in the $5 \times 5$ matrix form:
\begin{equation}
\label{representation}
    \mathbf{X}_{t} = \begin{bmatrix} \mathbf{R}^{w}_{b_{t}} & ^{w}\mathbf{v}_{b_{t}} & \mathbf{p}^{w}_{b_{t}} \\ 
    \mathbf{0}_{1 \times 3} & 1 & 0 \\ 
    \mathbf{0}_{1 \times 3} & 0 & 1
    \end{bmatrix}
\end{equation}
We use the right invariant error $\bm{\eta}_{t}^{r}$ \cite{barrau2016invariant} with elements lying in Lie algebra $\mathfrak{se}_{2}(3)$ which is formulated by first-order approximation as below: 
\begin{equation}
\label{SE3}
    \bm{\eta}_{t}^{r} = \hat{\mathbf{X}}_{t} \mathbf{X}_{t}^{-1} \approx \bm{I} + \bm{\xi}_{t}^{\wedge} = \begin{bmatrix} \mathbf{I} + (\bm{\xi}_{R_{t}})_{\times} & \bm{\xi}_{v_{t}} & \bm{\xi}_{p_{t}} \\ 
    \mathbf{0}_{1 \times 3} & 1 & 0 \\ 
    \mathbf{0}_{1 \times 3} & 0 & 1 
    \end{bmatrix} 
\end{equation}
where $(\cdot)^{\wedge}$ maps the Euclidean tangent space of Lie group to the corresponding Lie algebra space at the identity and $(\cdot)_{\times}$ denotes the $3\times 3$ skew-symmetric matrix. $\bm{I}$ is the group identity and $\mathbf{I}$ is the $3 \times 3$ identity matrix. We then have:
\begin{align}
    \begin{aligned}
        &\mathbf{I} + (\bm{\xi}_{R_{t}})_{\times} \approx \hat{\mathbf{R}}_{b_{t}}^{w} \mathbf{R}_{b_{t}}^{w^{\top}} \\ 
        &\bm{\xi}_{v_{t}} = {^{w}}\hat{\mathbf{v}}_{b_{t}} - \hat{\mathbf{R}}_{b_{t}}^{w} \mathbf{R}_{b_{t}}^{w^{\top}} {^{w}}\mathbf{v}_{b_{t}}\\
        &\bm{\xi}_{p_{t}} = \hat{\mathbf{p}}_{b_{t}}^{w} - \hat{\mathbf{R}}_{b_{t}}^{w} \mathbf{R}_{b_{t}}^{w^{\top}} \mathbf{p}_{b_{t}}^{w} 
    \end{aligned}
\end{align}


It is unfortunate that IMU biases can not fit into a Lie group. Therefore, we adopt the design of state representation in \cite{barrau2015non, potokar2021invariant} which treats the biases in a similar way as EKF. 
We use $\mathbf{b}_{t} = \begin{bmatrix} \mathbf{b}_{\omega_{t}}^{\top} & \mathbf{b}_{a_{t}}^{\top}
\end{bmatrix}^{\top} \in \mathbb{R}^{6}$ to denote the bias states and the corresponding bias error is defined as 
\begin{equation} 
\label{bias_error}
    \bm{\zeta}_{t} = {[\bm{\zeta}_{\omega_{t}}^{\top} \ \bm{\zeta}_{a_{t}}^{\top}]}^{\top} = {[(\hat{\mathbf{b}}_{\omega_{t}} - \mathbf{b}_{\omega_{t}})^{\top} \ (\hat{\mathbf{b}}_{a_{t}} - \mathbf{b}_{a_{t}})^{\top}]}^{\top}
\end{equation}

The system state can be propagated by IMU kinematic propagation model \cite{hartley2020contact} and the observation model for the correction step is given as
\begin{equation}
\label{measurement}
    \mathbf{y}_{t} = h(\mathbf{X}_{t}) + \mathbf{v}_{t}
\end{equation}
where $\mathbf{y}_{t}$ and $\mathbf{v}_{t}$ are the measurement vector and Gaussian noise vector, respectively. $h(\cdot)$ represents the observation model. To update the state with a new measurement, we use the form $\mathbf{z}_{t} = \mathbf{y}_{t} - h(\hat{\mathbf{X}}_{t})$ as measurement error. Substituting (\ref{measurement}) into the error function and expanding it in a first-order Taylor series approximation, we can get:
\begin{equation}
\label{linearized}
    \mathbf{z}_{t} = -\mathbf{H}_{t} \bm{\xi}_{t} + \mathbf{v}_{t}
\end{equation}
where $\mathbf{H}_{t}$ is the observation Jacobian matrix. From the linearized observation model (\ref{linearized}), we can derive the Kalman gain $\mathbf{K}_{t}$. Meanwhile, the state estimate and the covariance matrix $\mathbf{P}_{t}$ can be updated to posterior values. 

\subsection{IMU Propagation Model}
The IMU measurements corrupted by the accelerometer bias, gyroscope bias, and additive Gaussian white noises are
\begin{equation} 
    \Tilde{\bm{\omega}}_{t} = \bm{\omega}_{t} + \mathbf{b}_{\omega_{t}} + \mathbf{n}_{\omega_{t}}, \ \Tilde{\bm{a}}_{t} = \bm{a}_{t} + \mathbf{b}_{a_{t}} + \mathbf{n}_{a_{t}} 
\end{equation}
where $\Tilde{\bm{\omega}}_{t}$ and $\Tilde{\bm{a}}_{t}$ represent measurements of angular velocity and specific force in the IMU frame (i.e. the body frame $b$). $\mathbf{n}_{\omega_{t}} \sim \mathcal{N}(\mathbf{0}_{3 \times 1}, \mathbf{\Sigma}_{\omega})$ and $\mathbf{n}_{a_{t}} \sim \mathcal{N}(\mathbf{0}_{3 \times 1}, \mathbf{\Sigma}_{a})$ are the noises. The state dynamics model can be derived with the IMU propagation model \cite{potokar2021invariant} in continuous-time as
\begin{align}
\label{state}
\dot{\mathbf{R}}_{b_{t}}^{w} &= \mathbf{R}_{b_{t}}^{w} (\Tilde{\bm{\omega}}_{t} - \mathbf{b}_{\omega_{t}} - \mathbf{n}_{\omega_{t}})_{\times} \notag \\ 
{^{w}} \dot{\mathbf{v}}_{b_{t}} &= \mathbf{R}_{b_{t}}^{w}(\Tilde{\bm{a}}_{t} - \mathbf{b}_{a_{t}} - \mathbf{n}_{a_{t}}) + \mathbf{g}^{w} \notag \\ 
\dot{\mathbf{p}}_{b_{t}}^{w} &= {^{w}} \mathbf{v}_{b_{t}} \notag\\ 
\dot{\mathbf{b}}_{\omega_{t}} &= \mathbf{n}_{b\omega_{t}}, \ \dot{\mathbf{b}}_{a_{t}} = \mathbf{n}_{ba_{t}}
\end{align}
The biases are modeled as random walks, and thus the derivatives of biases follow zero-mean Gaussian distributions, i.e., $\mathbf{n}_{b\omega_{t}} \sim \mathcal{N}(\mathbf{0}_{3 \times 1}, \mathbf{\Sigma}_{b\omega})$, $\mathbf{n}_{ba_{t}} \sim \mathcal{N}(\mathbf{0}_{3 \times 1}, \mathbf{\Sigma}_{ba})$. 

For the states fitting into matrix Lie groups, by carrying out chain rules and the first-order approximation: $\bm{\eta}_{t}^{r} = \exp(\bm{\xi}_{t}) \approx \bm{I} + \bm{\xi}_{t}^{\wedge}$, the linearized model of right invariant error can be derived as
\begin{equation} 
\label{system_dynamics}
\setlength{\arraycolsep}{1.0pt}
	\begin{split}
            &\frac{d}{dt} \! \begin{bmatrix} \bm{\xi}_{R_{t}} \\ \bm{\xi}_{v_{t}} \\ \bm{\xi}_{p_{t}} \\ \bm{\zeta}_{\omega_{t}} \\ \bm{\zeta}_{a_{t}} \end{bmatrix} \! = \! \begin{bmatrix} \mathbf{0} & \mathbf{0} & \mathbf{0} & -\hat{\mathbf{R}}_{b_{t}}^{w} & \mathbf{0} \\ (\mathbf{g}^{w})_{\times} & \mathbf{0} & \mathbf{0} & -({^{w}}\hat{\mathbf{v}}_{b_{t}})_{\times} \hat{\mathbf{R}}_{b_{t}}^{w} & -\hat{\mathbf{R}}_{b_{t}}^{w} \\ \mathbf{0} & \mathbf{I} & \mathbf{0} & -(\hat{\mathbf{p}}_{b_{t}}^{w})_{\times} \hat{\mathbf{R}}_{b_{t}}^{w} & \mathbf{0} \\ \mathbf{0} & \mathbf{0} & \mathbf{0} & \mathbf{0} & \mathbf{0} \\ \mathbf{0} & \mathbf{0} & \mathbf{0} & \mathbf{0} & \mathbf{0} \end{bmatrix} 
            \begin{bmatrix} \bm{\xi}_{R_{t}} \\ \bm{\xi}_{v_{t}} \\ \bm{\xi}_{p_{t}} \\ \bm{\zeta}_{\omega_{t}} \\ \bm{\zeta}_{a_{t}} \end{bmatrix}+ \\ 
            &\! \begin{bmatrix} \hat{\mathbf{R}}_{b_{t}}^{w} & \mathbf{0} & \mathbf{0} & \mathbf{0} & \mathbf{0} \\ ({^{w}}\hat{\mathbf{v}}_{b_{t}})_{\times} \hat{\mathbf{R}}_{b_{t}}^{w} & \hat{\mathbf{R}}_{b_{t}}^{w} & \mathbf{0} & \mathbf{0} & \mathbf{0} \\ (\hat{\mathbf{p}}_{b_{t}}^{w})_{\times} \hat{\mathbf{R}}_{b_{t}}^{w} & \mathbf{0} & \hat{\mathbf{R}}_{b_{t}}^{w} & \mathbf{0} & \mathbf{0} \\ \mathbf{0} & \mathbf{0} & \mathbf{0} & \mathbf{I} & \mathbf{0} \\ \mathbf{0} & \mathbf{0} & \mathbf{0} & \mathbf{0} & \mathbf{I} \end{bmatrix}\! \begin{bmatrix} \mathbf{n}_{\omega_{t}} \\ \mathbf{n}_{a_{t}} \\ \mathbf{0}_{3 \times 1} \\ \mathbf{n}_{b \omega_{t}} \\ \mathbf{n}_{b a_{t}} \end{bmatrix}
            \!\triangleq\! \mathbf{A}_{t}\! \begin{bmatrix} \bm{\xi}_{t} \\ \bm{\zeta}_{t} \end{bmatrix}\! +\! \mathbf{Ad}_{\hat{\mathbf{X}}_{t}, \hat{\mathbf{b}}_{t}} \mathbf{n}_{t}
	\end{split}
\end{equation}
where $\mathbf{Ad}_{\hat{\mathbf{X}}_{t}, \hat{\mathbf{b}}_{t}}\triangleq block\_diag(\mathbf{Ad}_{\hat{\mathbf{X}}_{t}}, \mathbf{I}_{6 \times 6})$ and $\mathbf{Ad}_{\hat{\mathbf{X}}_{t}}$ is the adjoint matrix of $\hat{\mathbf{X}}_{t}$ \cite{hartley2020contact}. $\mathbf{0}$ is the $3 \times 3$ zero matrix. The supplementary material\cite{supplementary_materials} gives the detailed derivation. With the above continuous-time propagation model and (\ref{state}), the discretized system dynamics can be derived using Euler integration \cite{potokar2021invariant}:
\begin{align}
\label{discrete_dynamics} 
    \hat{\mathbf{R}}_{b_{t + \Delta t}}^{w-} &= \hat{\mathbf{R}}_{b_{t}}^{w+} \exp((\Tilde{\bm{\omega}}_{t} - \hat{\mathbf{b}}^{+}_{\omega_{t}}) \Delta t) \notag \\ 
    {^{w}}\hat{\mathbf{v}}_{b_{t + \Delta t}}^{-} &= {^{w}}\hat{\mathbf{v}}_{b_{t}}^{+} + (\hat{\mathbf{R}}_{b_{t}}^{w+}(\Tilde{\bm{a}}_{t} - \hat{\mathbf{b}}^{+}_{a_{t}}) + \mathbf{g}^{w}) \Delta t \notag \\ 
    \hat{\mathbf{p}}_{b_{t + \Delta t}}^{w-} &= \hat{\mathbf{p}}_{b_{t}}^{w+} + {^{w}}\hat{\mathbf{v}}_{b_{t}}^{+} \Delta t + \frac{1}{2}(\hat{\mathbf{R}}_{b_{t}}^{w+}(\Tilde{\bm{a}}_{t} - \hat{\mathbf{b}}_{a_{t}}^{+}) + \mathbf{g}^{w}) \Delta t^{2} \notag \\ 
    \hat{\mathbf{b}}^{-}_{\omega_{t + \Delta t}} &= \hat{\mathbf{b}}^{+}_{\omega_{t}}, \quad \hat{\mathbf{b}}^{-}_{a_{t + \Delta t}} = \hat{\mathbf{b}}^{+}_{a_{t}}  
\end{align}
where $\Delta t$ is the time step between two consecutive IMU samples. $(\cdot)^{-}$ and $(\cdot)^{+}$ represent the predicted and the updated state. The IMU measurements are assumed to be constant between adjacent samplings. The covariance is computed according to the Ricatti equation \cite{maybeck1982stochastic}:
\begin{equation}
\label{predicted_covariance} 
    \hat{\mathbf{P}}_{t + \Delta t}^{-} = \mathbf{\Phi}_{t} \hat{\mathbf{P}}_{t}^{+} \mathbf{\Phi}_{t}^{\top} + \mathbf{\Phi}_{t} \mathbf{Ad}_{\hat{\mathbf{X}}_{t}, \hat{\mathbf{b}}_{t}} \mathbf{Q}_{t} \mathbf{Ad}_{\hat{\mathbf{X}}_{t}, \hat{\mathbf{b}}_{t}}^{\top} \mathbf{\Phi}_{t}^{\top} \Delta t
\end{equation}
where $\mathbf{\Phi}_{t} = \exp_{m}(\mathbf{A}_{t} \Delta t)$ is the state transition matrix, $\exp_{m}(\cdot)$ is the exponential map of matrices. $\hat{\mathbf{P}}_{t}^{+}$ is the posterior covariance of $\bm{\xi}_{t}$ at time $t$. $\mathbf{Q}_{t} = block\_diag(\mathbf{\Sigma}_{\omega}, \mathbf{\Sigma}_{a}, \mathbf{0}, \mathbf{\Sigma}_{b \omega}, \mathbf{\Sigma}_{b a})$ is the covariance of  $\mathbf{n}_{t}$. 

\subsection{Odometer Observation Model}
The wheel odometer with encoders can provide linear velocity measurements for wheel robots. The measured velocity is also assumed to be corrupted by Gaussian white noises
\begin{equation}
\label{noisy_velocity} 
    {^{o}}\Tilde{\mathbf{v}}_{o_{t}} = {^{o}}\mathbf{v}_{o_{t}} + \mathbf{n}_{o_{t}}^{\prime}
\end{equation}
where $\mathbf{n}_{o_{t}}^{\prime} \sim \mathcal{N}(\mathbf{0}_{3 \times 1}, \mathbf{\Sigma}_{o}^{\prime})$. The velocity expressed in the IMU frame ${^{b}}\Tilde{\mathbf{v}}_{b_{t}}$ and its covariance can be transformed through the calibrated extrinsic parameters $(\mathbf{R}^{b}_{o}, \mathbf{t}^{b}_{o})$ between the odometer and body (IMU) frame
\begin{gather}
\label{transformed_velocity} 
    {^{b}}\Tilde{\mathbf{v}}_{b_{t}} = \mathbf{R}_{o}^{b}{^{o}}\Tilde{\mathbf{v}}_{o_{t}} + (\mathbf{t}_{o}^{b})_{\times} (\Tilde{\bm{\omega}}_{t} - \mathbf{b}_{\omega_{t}}) \\ 
    \mathbf{\Sigma}_{o} = \mathbf{R}_{o}^{b} \mathbf{\Sigma}_{o}^{\prime} \mathbf{R}_{o}^{b^{\top}} + (\mathbf{t}_{o}^{b})_{\times} (\mathbf{\Sigma}_{\omega} + \mathbf{\Sigma}_{b\omega})(\mathbf{t}_{o}^{b})_{\times}^{\top}
\end{gather}
In our case, the IMU and odometer are assumed to reside in the same location with only $\mathbf{R}_{o}^{b}$ being considered. The observation model under the right invariant form is
\begin{equation} 
\label{odometer_observation} 
    \mathbf{y}_{t} = {^{b}}\Tilde{\mathbf{v}}_{b_{t}} = h(\mathbf{X}_{t}) + \mathbf{n}_{o_{t}}, \quad h(\mathbf{X}_{t}) = \mathbf{R}_{b_{t}}^{w^{\top}}{^{w}} \mathbf{v}_{b_{t}} 
\end{equation} 
where $\mathbf{n}_{o_{t}} \sim \mathcal{N}(\mathbf{0}_{3 \times 1}, \mathbf{\Sigma}_{o})$. As in (\ref{linearized}), the measurement error can be derived as
\begin{flalign}
\label{odometer_measurement_error} 
    \mathbf{z}_{o_{t}}\!\kern-2pt =\!\kern-2pt -\!\kern-2pt \begin{bmatrix} \mathbf{0}\!\kern-4pt & \mathbf{I}\!\kern-4pt & \mathbf{0}\!\kern-4pt & \mathbf{0}\!\kern-4pt & \mathbf{0} \end{bmatrix}\!\kern-2pt \begin{bmatrix} \bm{\xi}_{t} \\ \bm{\zeta}_{t} \end{bmatrix}\!\kern-2pt +\!\kern-2pt \hat{\mathbf{R}}_{b_{t}}^{w} \mathbf{n}_{o_{t}}\!\kern-2pt \triangleq\!\kern-2pt -\mathbf{H}_{o_{t}}\!\kern-2pt \begin{bmatrix} \bm{\xi}_{t} \\ \bm{\zeta}_{t} \end{bmatrix}\!\kern-2pt +\!\kern-2pt \hat{\mathbf{R}}_{b_{t}}^{w} \mathbf{n}_{o_{t}}
\end{flalign}
Using the Kalman filtering theory\cite{barrau2016invariant}, the updated states and covariance can be obtained.

\subsection{Camera Observation Model}
%

\subsubsection{Data Association} Assuming there are $n_{t}$ detected streetlights and $m_{t}$ 3D streetlight clusters within a certain range of $th_{Lr1}$ from the robot pose estimate $\hat{\mathbf{R}}_{b_{t}}^{w}, \hat{\mathbf{p}}_{b_{t}}^{w}$. The streetlight observations are represented as $\{B_{t_{i}}=(\Tilde{\mathbf{p}}_{t_{i}},\Tilde{\bm{l}}_{t_{i}})\}_{i=1}^{n_{t}}$ where $\Tilde{\mathbf{p}}_{t_{i}}=\begin{bmatrix}
\Tilde{u}_{t_{i}}&\Tilde{v}_{t_{i}}
\end{bmatrix}^{\top}$ is the image coordinate of the center of object detection bounding boxes. $\Tilde{\bm{l}}_{t_{i}}=\begin{bmatrix}
\Tilde{lu}_{t_{i}} & \Tilde{lv}_{t_{i}}
\end{bmatrix}^{\top}$ contains the width and length of box. The streetlight clusters in the map are represented as $\{L_{j}=(\Tilde{\mathbf{C}}_{j}^{w},{^{1}}\Tilde{\mathbf{P}}^{w}_{j},...,{^{J}}\Tilde{\mathbf{P}}^{w}_{j}),(-1)\}_{j=1}^{m_{t}+1}$ where $\Tilde{\mathbf{C}}_{j}^{w}=\frac{1}{J}\Sigma^{J}_{jj=1}{^{jj}}\Tilde{\mathbf{P}}^{w}_{j}$ is the geometric center and ${^{jj}}\Tilde{\mathbf{P}}^{w}_{j}\in \mathbb{R}^{3}$ denotes the 3D coordinate of each point. Since there may be false detections of streetlights, the set of streetlight clusters is extended with element $(-1)$. We score all combinations of matches between 2D and 3D data and the one with the highest score is determined as the correct data association. The score is calculated based on two kinds of match errors.

\textbf{Reprojection Error}. Given a prior pose, an incorrect match generally results in a large error between the projected center of the 3D streetlight cluster and the box center. We model the residual vector $\mathbf{r}_{t_{proj}}$ as Gaussian distribution. Then for detected box $B_{t_{i}}$ and streetlight cluster $L_{j}$, the reprojection error-based score $s_{t_{proj,ij}}$ is calculated as
\begin{align}
\label{reprojection_score} 
    s_{t_{proj, ij}} &= \mathcal{N}(\mathbf{r}_{t_{proj, ij}} | \mathbf{0}_{3 \times 1}, \mathbf{\Sigma_{t_{proj, ij}}}) \notag \\ 
    \mathbf{r}_{t_{proj, ij}} &= \Tilde{\mathbf{p}}_{t_{i}} - \pi_{c}(\mathbf{R}^{c}_{b} \hat{\mathbf{R}}^{w^{\top}}_{b_{t}}(\Tilde{\mathbf{C}}^{w}_{j} - \hat{\mathbf{p}}^{w}_{b_{t}}) + \mathbf{t}_{b}^{c})
\end{align}
where $\mathcal{N}(\cdot | \mathbf{0}_{3 \times 1}, \mathbf{\Sigma_{t_{proj, ij}}})$ represents the Gaussian probability density function with mean $\mathbf{0}_{3 \times 1}$ and covariance $\mathbf{\Sigma_{t_{proj,ij}}}$ in the 3D Euclidean space. $\mathbf{R}^{c}_{b}$ and $\mathbf{t}_{b}^{c}$ constitute the extrinsics between IMU and camera. $\pi_{c}(\cdot)$ is the perspective projection function of the camera. 
	
\textbf{Angle Error}. We also introduce the score based on angle error $\theta_{ij}$ between the back-projection ray of the center of box $B_{t_{i}}$ and the line connecting the center of streetlight cluster $L_{j}$ and the optical center of the camera, which is defined as
\begin{align}
\label{angle_error} 
    s_{t_{ang, ij}} &= \mathcal{N}(r_{t_{ang,ij}} | 0, \sigma^{2}_{t_{ang, ij}}) \notag \\ 
    r_{t_{ang, ij}} &= 1 - \cos(\theta_{ij})
    = 1 - \frac{\Tilde{\mathbf{P}}_{t_{i}}^{\top} \Tilde{\mathbf{C}}_{j}^{c}}{\lVert\Tilde{\mathbf{P}}_{t_{i}}\rVert_{2}\cdot\lVert\Tilde{\mathbf{C}}_{j}^{c}\rVert_{2}}
\end{align}
where $\Tilde{\mathbf{P}}_{t_{i}} = \pi_{c}^{-1}(\Tilde{\mathbf{p}}_{t_{i}})$ is the back-projection of box center. The derivations of $\mathbf{\Sigma_{t_{proj, ij}}}$ and $\sigma^{2}_{t_{ang, ij}}$ are provided in the supplementary material.

Then the total score of each match is formulated as
\begin{align}
	s_{ij}=\omega s_{t_{proj,ij}}+(1-\omega)s_{t_{ang,ij}}
\end{align}
where $0 \leq \omega \leq 1$ is the preset weight. For the streetlight $i$ without any correspondence, the score is calculated as
\begin{equation}
\label{no_match_score}
	s_{ij} = 1 - \sum_{m = 1}^{m_{t}} s_{im}, \quad j > m_{t}
\end{equation}
We use $d_{ij}$ to indicate whether the $i$-th streetlight observation matches the $j$-th 3D streetlight cluster. $\mathcal{D}_{t} = \{((B_{t_{i}}, L_{j}) | d_{ij} = 1)\}_{i = 1}^{n_{t}}$ denotes the possible match combinations. Note that all streetlight observations have a one-to-one mapping with 3D clusters. We expand the set of clusters to $m_{t} + n_{t}$ with $n_{t}$ elements representing no correspondence of 2D observations so that this one-to-one mapping feature can be maintained when observation $B_{i}$ is mapped to $(-1)$. To obtain the best combination $\mathcal{D}_{t}^{\star}$, the optimization problem is formulated as the classical assignment problem
\begin{align}
\label{assignment_problem} 
    \mathcal{D}_{t}^{\star}\! =\! \mathop{\arg\max} \limits_{\mathcal{D}_{t}}\! \sum_{i = 1}^{n_{t}}\! \sum_{j = 1}^{m_{t} + n_{t}}\! d_{ij} s_{ij} ,\ \text{s.t. } \sum_{i = 1}^{n_{t}}\! d_{ij}\! \leq\! 1 \ \sum_{j = 1}^{m_{t} + n_{t}}\! d_{ij}\! =\! 1
\end{align}
The optimization problem can be fast solved ($\sim$0.2 ms per frame) by Hungarian algorithm\cite{kuhn1955hungarian}. 

After obtaining the 2D-3D matches, the camera observation model can be derived. Only matches of $\mathcal{D}_{t}^{\star}$ without $L_{j} = (-1)$ are selected. For the $i$-th match $(B_{t_{i}},L_{j})$, the observation model is formulated as
\begin{align}
\label{camera-based observation model} 
    \mathbf{y}_{t_{i}} &= \pi_{c}^{-1}(\Tilde{\mathbf{p}}_{t_{i}}) = h_i(\mathbf{X}_{t}) + \mathbf{n}_{c} \notag \\ 
    h_i(\mathbf{X}_{t}) &= [\mathbf{R}_{b}^{c} \mathbf{R}_{b_{t}}^{w^{\top}}(\Tilde{\mathbf{C}}_{t_{j}}^{w} - \mathbf{p}^{w}_{b_{t}}) + \mathbf{t}_{b}^{c}] / {Z_{t_{j}}^{c}} 
\end{align}
where $\mathbf{n}_{c} \sim \mathcal{N}(\mathbf{0}, \mathbf{K}^{-1} \mathbf{\Sigma}_{c} \mathbf{K}^{-\top})$ and $\mathbf{K}$ is the camera intrinsics matrix. $\mathbf{\Sigma}_{c}$ is a $3 \times 3$ diagonal matrix with 0 as the last element. According to (\ref{linearized}), by defining $\hat{\mathbf{\Psi}}_{t_{i}} \triangleq \hat{\mathbf{R}}_{b_{t}}^{w^{\top}}(\Tilde{\mathbf{C}}_{t_{j}}^{w} - \hat{\mathbf{p}}^{w}_{b_{t}})$, the observation matrix can be derived as
\begin{equation} 
\label{camera_measurement_error_i} 
    \setlength{\arraycolsep}{1.1pt} 
    \begin{aligned} 
        \mathbf{z}_{c_{t_{i}}} &= -\mathbf{H}_{i}(\hat{\mathbf{\Psi}}_{t_{i}}) \hat{\mathbf{R}}_{b_{t}}^{w^{\top}} \begin{bmatrix} (\Tilde{\mathbf{C}}_{t_{j}}^{w})_{\times} & \mathbf{0} & -\mathbf{I} & \mathbf{0} & \mathbf{0} \end{bmatrix} \begin{bmatrix} \bm{\xi}_{t} \\ \bm{\zeta}_{t} \end{bmatrix} + \mathbf{n}_{c} \\ 
        & \triangleq - \mathbf{H}_{c_{t_{i}}} \begin{bmatrix} \bm{\xi}_{t} \\ \bm{\zeta}_{t} \end{bmatrix} + \mathbf{n}_{c} 
    \end{aligned}
\end{equation}
where $\mathbf{H}_{i}(\hat{\mathbf{\Psi}}_{t_{i}}) \triangleq \frac{\partial h_{i}(\mathbf{\Psi}_{t_{i}})}{\partial \mathbf{\Psi}_{t_{i}}} \Big|_{\mathbf{\Psi}_{t_{i}} = \hat{\mathbf{\Psi}}_{t_{i}}}$. The detailed derivation is given in the supplementary material.
The measurement error is obtained as
\begin{flalign}
\label{camera_measurement_error} 
    \mathbf{z}_{c_{t}}\kern-3pt\! =\!\kern-3pt \begin{bmatrix} \mathbf{z}_{c_{t_{1}}} \\ 
    \vdots \\ 
    \mathbf{z}_{c_{t_{n_{t}}}} \end{bmatrix}\kern-3pt\! =\!\kern-3pt \begin{bmatrix} \mathbf{H}_{c_{t_{1}}} \\ 
    \vdots \\ 
    \mathbf{H}_{c_{t_{n_{t}}}} \end{bmatrix}\!\! \begin{bmatrix} \bm{\xi}_{t} \\ \bm{\zeta}_{t} \end{bmatrix}\kern-2pt\! +\!\kern-2pt \begin{bmatrix} \mathbf{n}_{c} \\ 
    \vdots \\ 
    \mathbf{n}_{c} \end{bmatrix}\! \triangleq\! \mathbf{H}_{c_{t}}\! \begin{bmatrix} \bm{\xi}_{t} \\ \bm{\zeta}_{t} \end{bmatrix}\! +\! \mathbf{\Pi} \mathbf{n}_{c} 
\end{flalign}
where $\mathbf{\Pi} = \begin{bmatrix} \mathbf{I} & \cdots & \mathbf{I} \end{bmatrix}^{\top} \in \mathbb{R}^{3 n_{t} \times 3}$.

\begin{figure}[t] 
    \centering 
    \vspace{-0.3cm}
    \includegraphics[width=0.48\textwidth]{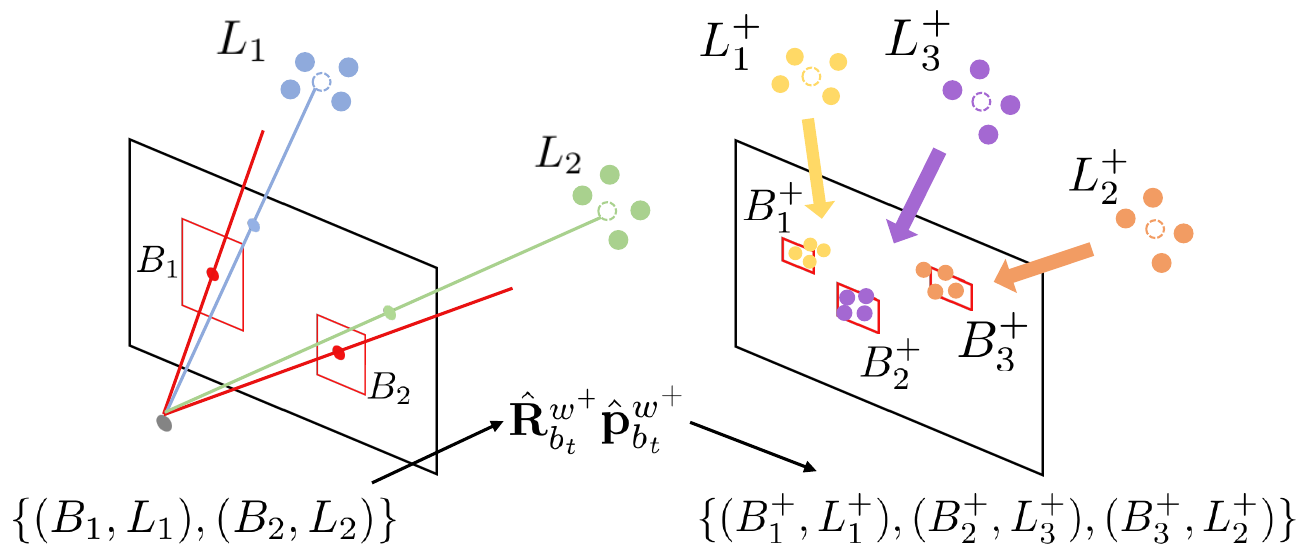} 
    \caption{An example of match extension. The module determines more matches according to the positional relationship between the projected locations of streetlight clusters and the boxes detected by the binary segmentation method.} 
    \label{match_extension}
    \vspace{0.2cm}
\end{figure}

\subsubsection{Match Extension} Despite the streetlight detector can identify the streetlights precisely, small lights are easily ignored, leading to a shortage of viable streetlight matches. To remedy this problem, we introduce the binary segmentation-based streetlight detection method. Fig. \ref{match_extension} shows an example of match extension. Pixels with intensity larger than $th_{int}$ are fed into the Suzuki85\cite{suzuki1985topological} algorithm to generate multiple bounding boxes which are considered as streetlight observations. Then with the updated robot pose $\hat{\mathbf{R}}_{b_{t}}^{w+}$ and $\hat{\mathbf{p}}^{w+}_{b_{t}}$, all streetlight map clusters within a range of $th_{Lr2}$ and not having 2D observation matches are projected to the image. 
To take one streetlight cluster as an example, since the updated pose is relatively accurate, all boxes containing the projected points are considered as the candidate matches. For each box, we then calculate the ratio of the number of projected points falling into the box to the total number of points within the streetlight cluster. The box with the maximum ratio is determined as the match of the cluster. By traversing all streetlight clusters, multiple new matches are selected and the robot state can be updated again. 
\subsubsection{Environmental Degeneration} \label{degeneration section} 
\begin{figure}[htbp]
    \centering
    \vspace{-0.3cm}
    \includegraphics[width=0.47 \textwidth]{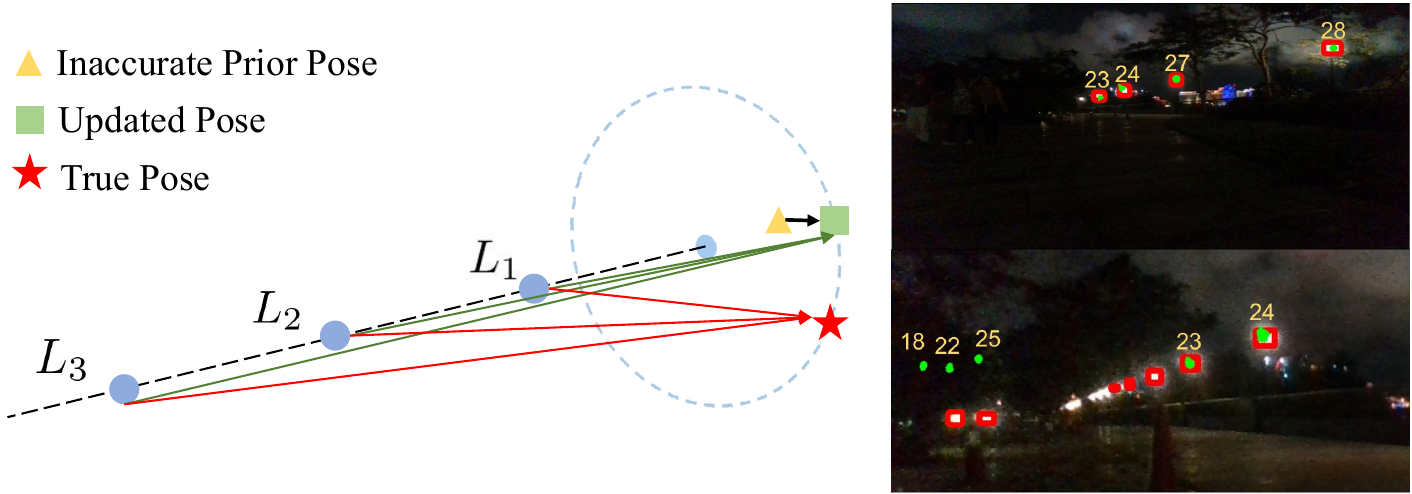}
    \caption{
    Degeneration case. The left image shows that all positions on the blue circle satisfy the identical camera observation constraints. An inaccurate prior pose leads to deviation from the true pose. The right images show that the streetlight matches are difficult to search after the degeneration cases.}
    \label{degeneration}
    \vspace{0.1cm}
\end{figure}

When the streetlights are in a straight line, as shown in Fig. \ref{degeneration}, any camera pose lying on the blue dotted circle satisfies the geometric constraints, thus the pose can drift with time.
After the robot experiences a period of degeneration, the matches of new streetlights that are not in a line with already observed streetlights are difficult to search as shown in Fig. \ref{degeneration}. We further find the positional drift in $z$-axis and rotational drifts in roll and pitch are more obvious (Section \ref{tracking_lost_and_degeneration_cases}). The reason is that the odometer can provide 2D motion measurements thus constraining the 3-DoF ($x$, $y$, and yaw). This case results in the projected points of the newly observed streetlights typically being located directly above or below the detection box. Using this feature, a direct and effective way is proposed. When the degeneration ends (i.e., new streetlights not in a line with already observed ones are detected), each projected center of newly observed streetlight clusters is assigned a rectangle. If the rectangle has an intersection area with detected boxes, the box closest to the projected center is chosen as the match of the corresponding streetlight cluster. Then the pose can be corrected by adding streetlights not in a straight line. Note that the height of the rectangle is significantly larger than its width due to the significant drifts in $z$-axis, roll, and pitch.

\subsection{Tracking Recovery}
In the cases without visible streetlights, the estimated trajectory could drift with time. The tracking recovery module is proposed to recover the pose estimate in the map. During tracking failures, brute-force matching is performed to traverse all match combinations. Assuming that there are $n_{re}$ detected boxes and $m_{re}$ streetlight clusters, each combination is assigned a punishment score calculated as:
\begin{equation}
    s_{re}\!=\!\frac{1}{n_{re}}\!\sum_{i = 1}^{n_{re}}\! \Delta p^{b_{re} \!+}_{i}\! + \!\gamma_{1}\Delta t_{b_{re}}^{b_{re} \!+}\! + \!\gamma_{2}\Delta\theta_{b_{re}}^{b_{re} \!+}\! + \!\gamma_{3}\!\sum_{i=1}^{n_{re}}\!neg(i)\notag
\end{equation}
\begin{align}
\label{reloc_score}
    \Delta p^{b_{re} +}_{i} &= \lVert\pi_{c}(\mathbf{R}_{b}^{c} \hat{\mathbf{R}}_{b_{re}}^{{w +}^{\top}}(\Tilde{\mathbf{C}}_{re_{j}}^{w} - \hat{\mathbf{p}}^{w +}_{b_{re}}) + \mathbf{t}_{b}^{c}) - \Tilde{\mathbf{p}}_{re_{i}}\rVert_{2} \notag \\ 
    \Delta t_{b_{re}}^{b_{re} +} &= \lVert\hat{\mathbf{p}}_{b_{re},1,2}^{w -} - \hat{\mathbf{p}}_{b_{re},1,2}^{w +}\rVert_{2} \notag \\ 
    \Delta \theta_{b_{re}}^{b_{re} +} &= yaw(\hat{\mathbf{R}}_{b_{re}}^{{w -}^{\top}}\hat{\mathbf{R}}_{b_{re}}^{w +})
\end{align}
where the updated pose $(\hat{\mathbf{R}}_{b_{re}}^{w+}, \hat{\mathbf{p}}^{w+}_{b_{re}})$ is calculated based on the combination. $neg(i) = 1$ when the $i$-th box has no correspondence, otherwise $neg(i) = 0$. $\gamma_{1}, \gamma_{2}, \gamma_{3}$ are utilized for balancing different terms. We only use the yaw angle of relative rotation ($yaw(\cdot)$) and the $x$, $y$ components of relative translation ($(\cdot)_{1, 2}$). Once there is a combination whose score is less than $th_{RE\_sc}$ and more than $2$ matches are found, it is considered to successfully relocalize the robot. Subsequently, the updated pose of the combination with the minimum score is regarded as the recovered pose. 

\section{Experiments}
\subsection{Implementation Details}
The robot is equipped with a front-facing RGB camera ($30$ Hz $1280 \times 720$ images), an odometer ($10$ Hz), and a Livox $360^{\circ}$ FOV LiDAR\footnote{https://www.livoxtech.com/mid-360} ($10$ Hz) integrating an ICM40609 IMU ($200$ Hz). The camera's intrinsics and all extrinsics among sensors have been pre-calibrated. All experiments were performed on the Intel i7-1165G7 CPU. Note that only an IMU, an odometer, and a monocular camera are used for the proposed nocturnal vision-aided localization system. 

We use Yolov5-Lite \cite{chen2021yolov5} as the streetlight detector for its real-time performance on CPU. 
The $2476$ images labeled from different nighttime scenes are used to train the network. The relatively accurate initial pose is manually given and is optimized by the PnP solver\cite{lepetit2009ep} with streetlight matches. 
\begin{figure}[t] 
    \centering
    \vspace{-0.5cm}
    \subfloat[Scene 1]{\includegraphics[width=0.225\textwidth]{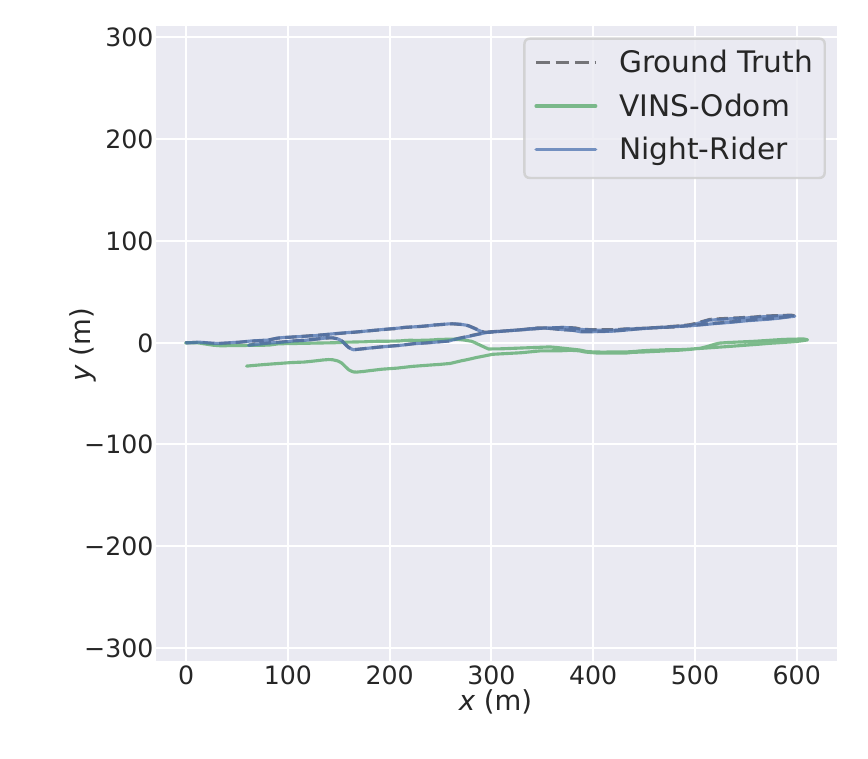}} 
    \hfill 
    \subfloat[Scene 2]{\includegraphics[width=0.225\textwidth]{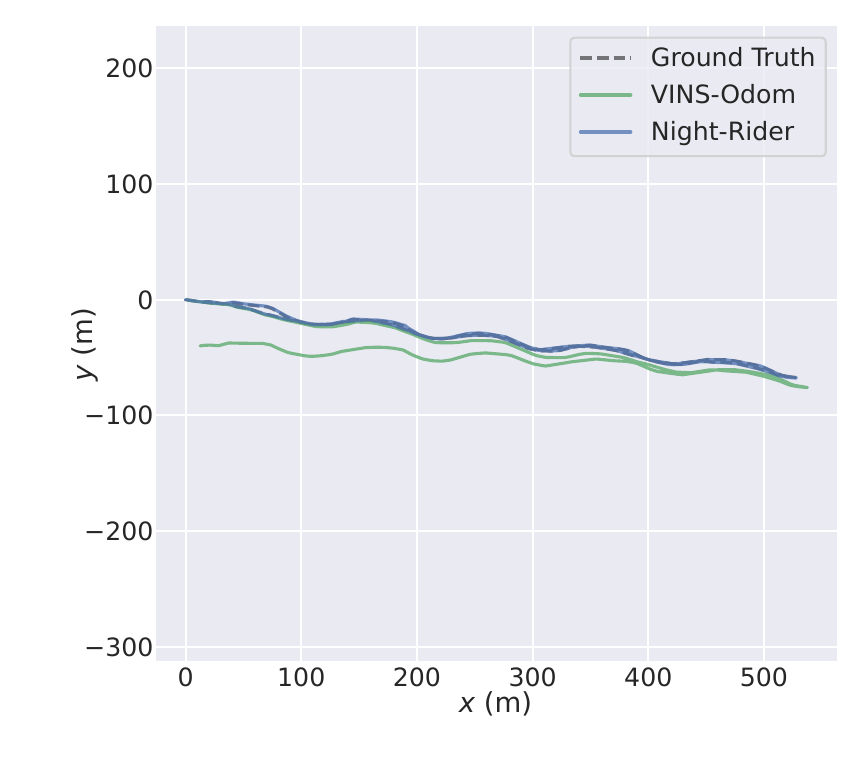}}
    \vspace{-2mm} 
    \newline 
    \subfloat[Scene 3]{\includegraphics[width=0.225\textwidth]{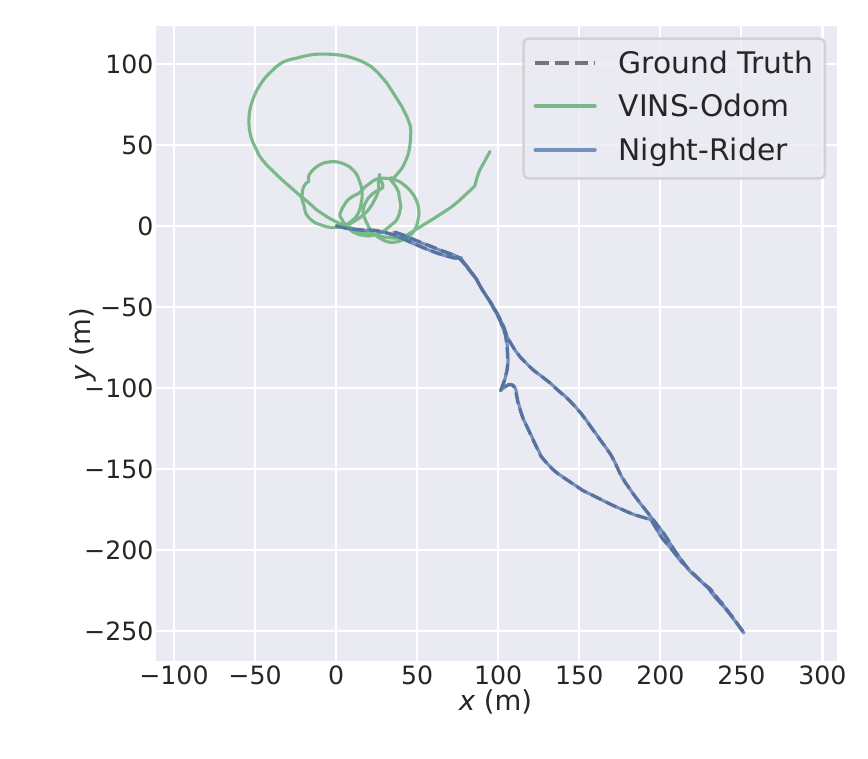}} 
    \hfill 
    \subfloat[Scene 4]{\includegraphics[width=0.225\textwidth]{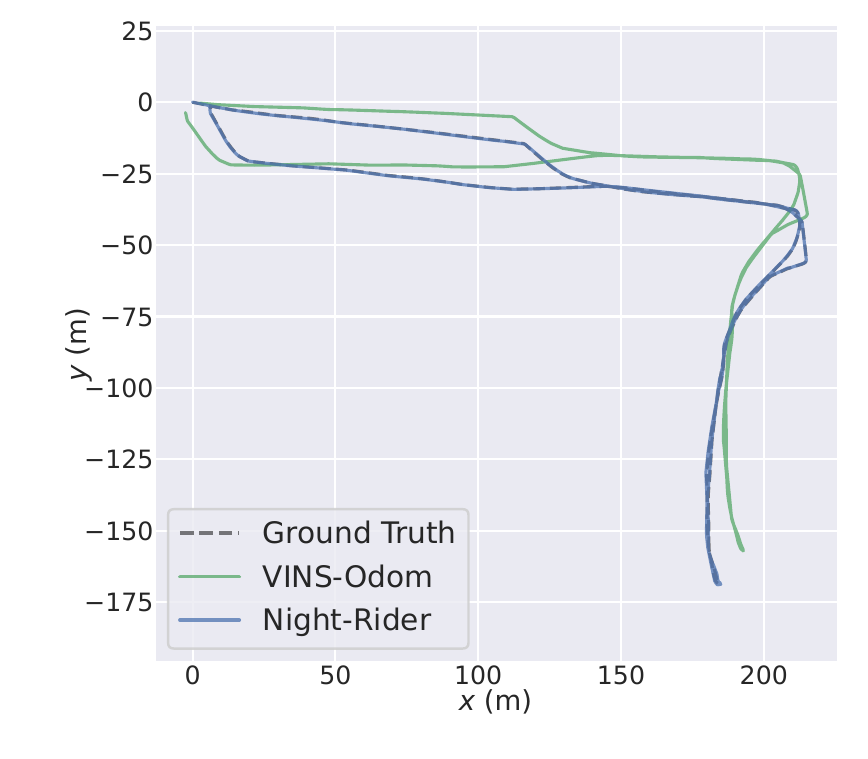}} 
    \caption{Trajectories of ground truth (\textcolor{gray}{gray} dashed line), VINS-Odom (\textcolor{green}{green}), and Night-Rider (\textcolor{blue}{blue}) in four real scenes.}
    \label{Comparison}
\end{figure}

\subsection{Evaluation of Localization}
To thoroughly evaluate the performance of the proposed system in nocturnal environments, we collect data in four different real scenes with multiple distribution types of streetlights. In the first scene (1149.2 m) and the fourth scene (723.8 m), the streetlights distribute sparsely on both sides of the road. Partial streetlights lie on only one side of the road in the second scene (1070.9 m). The third scene (777.2 m) contains densely distributed streetlights. All streetlight maps of these scenes are constructed using the method described in Section \ref{mapping}. The offline optimized and precisely aligned trajectories from the loop closure augmented FAST-LIO2 using the LiDAR on board are treated as ground truths and EVO \cite{grupp2017evo} is used for evaluation. Note that there are very few open-source research works targeting nocturnal scenarios. Thus we first compare Night-Rider with the state-of-the-art framework, extended VINS-Mono \cite{qin2018vins} with the odometer\cite{liu2019visual} (named VINS-Odom). As shown in Fig. \ref{Comparison}, Night-Rider can precisely estimate trajectories with extremely low drifts in all four scenes while VINS-Odom drifts over time and even fails to track in the third scene. The reason is that the unstable illumination in nighttime scenes makes it difficult to extract reliable feature points from highly bright or dark images. This instability results in drifts or tracking failures of VINS-Odom. Contrarily, Night-Rider is independent of feature points and exploits the stable streetlights for localization. Thus trajectories of Night-Rider align accurately with ground truths. Table. \ref{Table_Comparison} depicts the evaluation results of absolute trajectory error (ATE\cite{sturm2012benchmark}) for translation and rotation parts. It shows that Night-Rider achieves remarkably accurate and robust localization in all scenes (\textless 0.2\% relative error of total trajectory length) and totally outperforms VINS-Odom, which further validates the effectiveness of our system. 

\begin{figure}[t] 
    \centering
    \vspace{-0.5cm}
    \subfloat[Scene 1]{\includegraphics[width=0.24\textwidth]{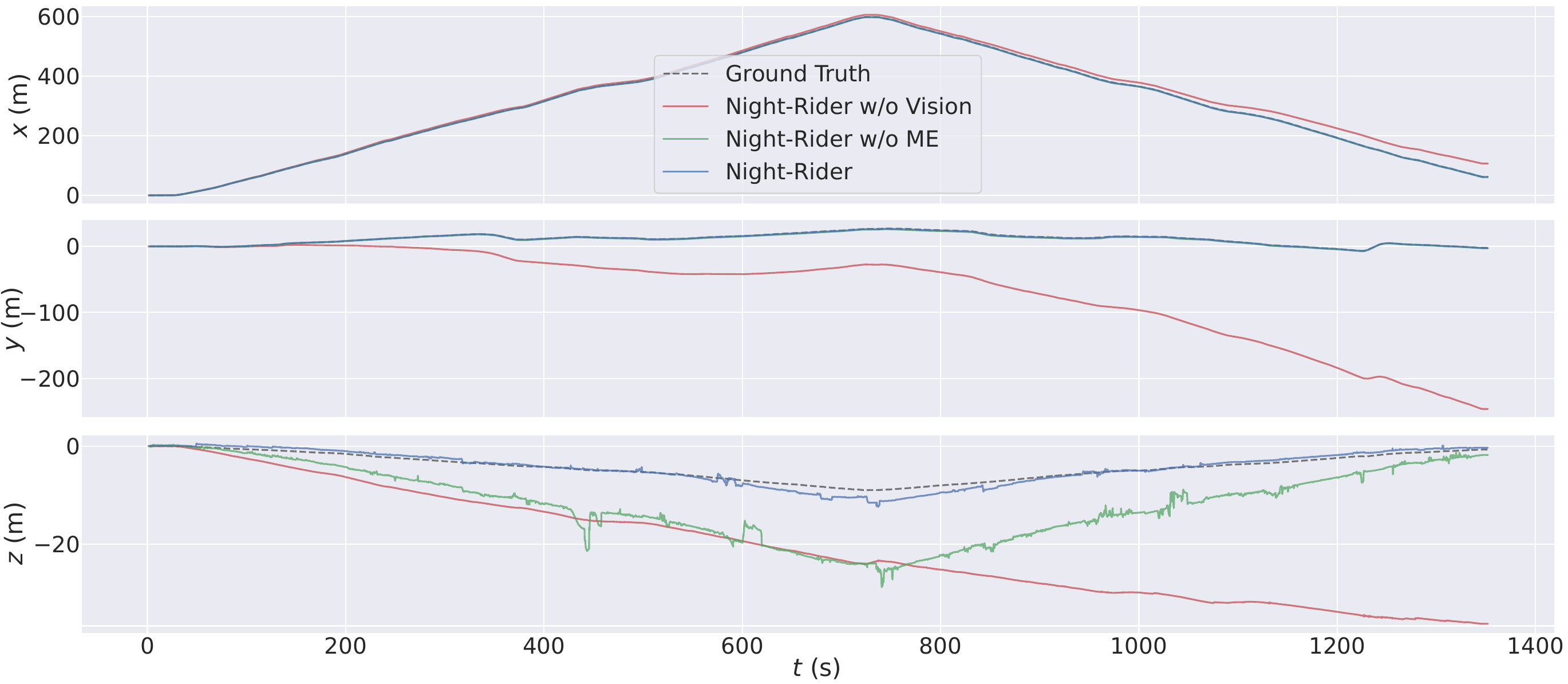}}
    \hfill
    \subfloat[Scene 2]{\includegraphics[width=0.24\textwidth]{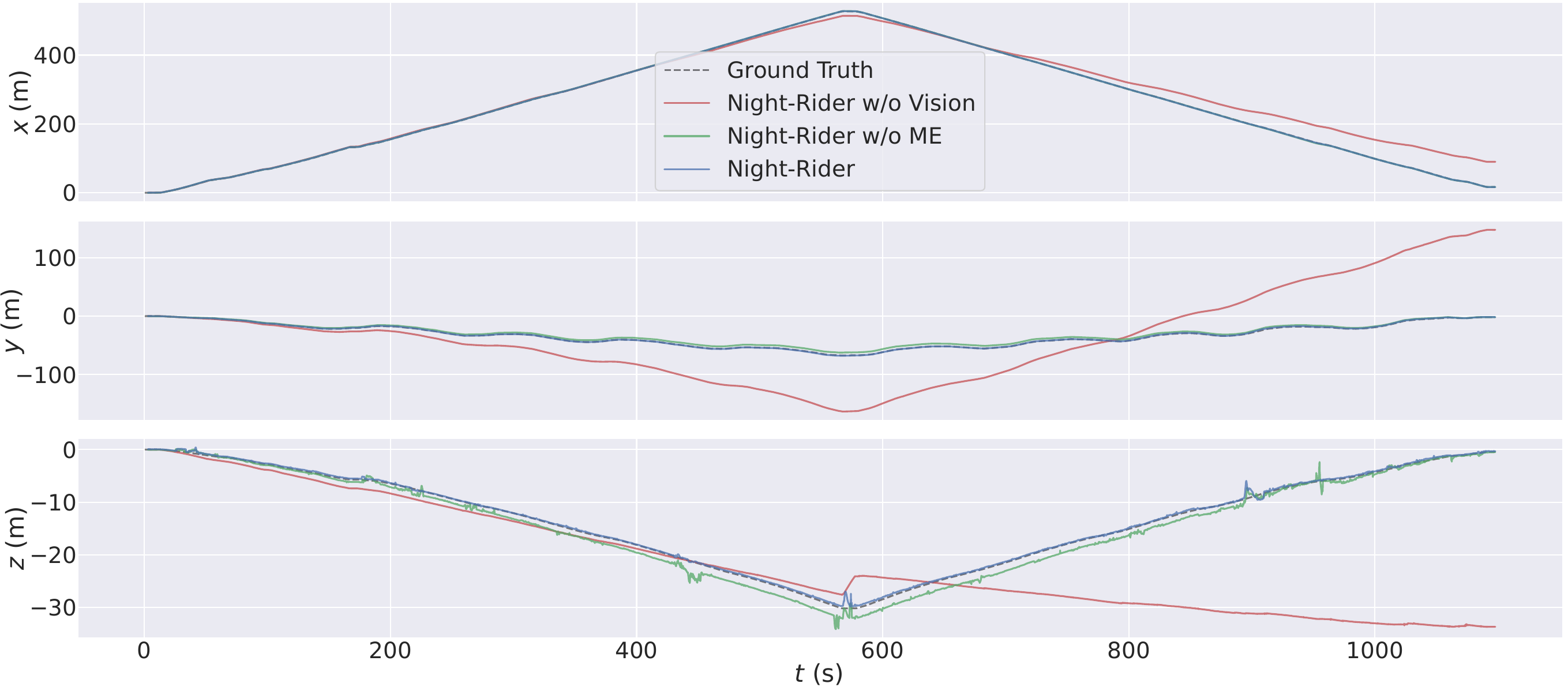}}
    \vspace{-0.3cm} 
    \newline 
    \subfloat[Scene 3]{\includegraphics[width=0.24\textwidth]{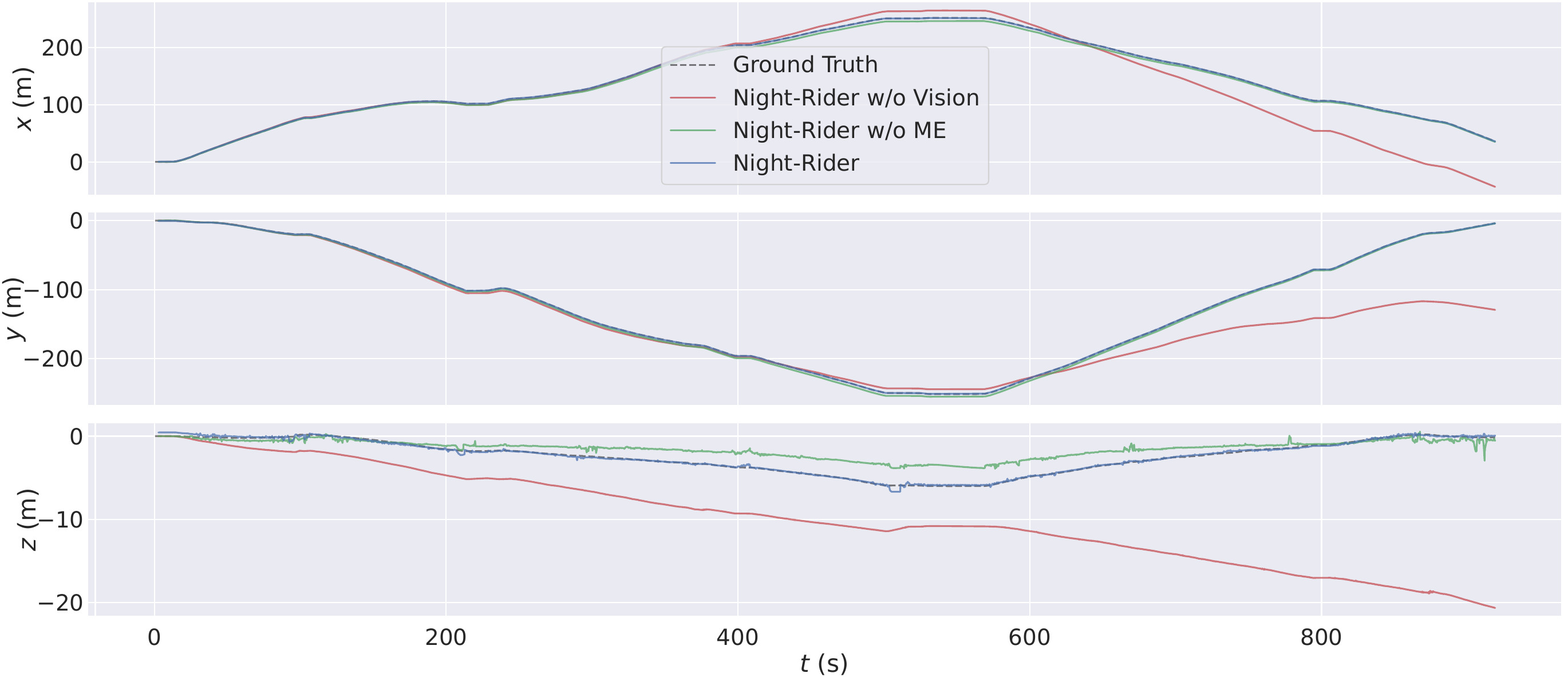}} 
    \hfill 
    \subfloat[Scene 4]{\includegraphics[width=0.24\textwidth]{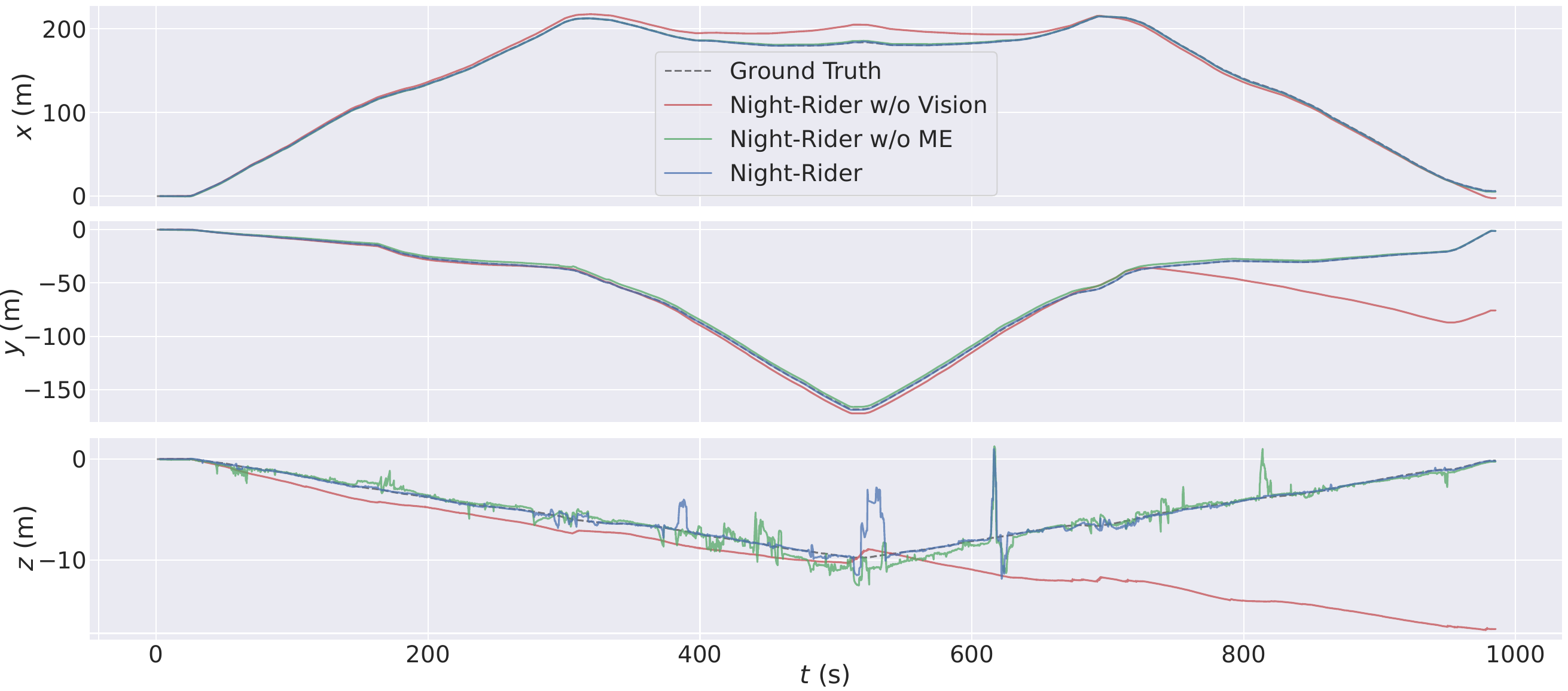}} 
    \caption{Trajectories in $x$, $y$ and $z$-axis of ground truth (\textcolor{gray}{gray} dashed line), Night-Rider w/o Vision (\textcolor{red}{red}), Night-Rider w/o ME (\textcolor{green}{green}), and Night-Rider (\textcolor{blue}{blue}) in four real scenes.} 
    \label{Ablation}
\end{figure}

\renewcommand 
\arraystretch{1.1}
\setlength{\belowcaptionskip}{0pt}
\begin{table}[htbp]
\setlength
\tabcolsep{2.2pt}
\centering
\vspace{-0.2cm}
\caption{ATE\cite{sturm2012benchmark} for translation and rotation of VINS-Odom, Night-Rider w/o Vision, Night-Rider w/o ME, and Night-Rider in four scenes. The best is shown in \textbf{bold}.} 
\begin{tabular}{ccclclclcl} 
\toprule
& & \multicolumn{2}{c}{VINS-Odom} & \multicolumn{2}{c}{\begin{tabular}[c]{@{}c@{}}Night-Rider\\w/o Vision\end{tabular}} & \multicolumn{2}{c}{\begin{tabular}[c]{@{}c@{}}Night-Rider\\w/o ME\end{tabular}} & \multicolumn{2}{c}{Night-Rider}\\
\midrule
\multirow{2}{*}{Scene 1} & Trans. (m) & \multicolumn{2}{c}{20.770} & \multicolumn{2}{c}{108.707} & \multicolumn{2}{c}{8.617} & \multicolumn{2}{c}{{\textbf{1.543}}}  \\
& Rot. (deg) & \multicolumn{2}{c}{2.959} & \multicolumn{2}{c}{15.553} & \multicolumn{2}{c}{4.071} & \multicolumn{2}{c}{{\textbf{2.307}}} \\
\midrule
\multirow{2}{*}{Scene 2} & Trans. (m) & \multicolumn{2}{c}{24.200} & \multicolumn{2}{c}{71.678} & \multicolumn{2}{c}{3.398} & \multicolumn{2}{c}{{\textbf{0.589}}} \\
& Rot. (deg) & \multicolumn{2}{c}{5.521} & \multicolumn{2}{c}{21.444} & \multicolumn{2}{c}{2.640} & \multicolumn{2}{c}{{\textbf{1.831}}} \\
\midrule
\multirow{2}{*}{Scene 3} & Trans. (m) & \multicolumn{2}{c}{221.422} & \multicolumn{2}{c}{57.864} & \multicolumn{2}{c}{4.072} & \multicolumn{2}{c}{{\textbf{0.339}}} \\
& Rot. (deg) & \multicolumn{2}{c}{109.834} & \multicolumn{2}{c}{19.841} & \multicolumn{2}{c}{2.554} &  \multicolumn{2}{c}{{\textbf{1.674}}}\\ 
\midrule
\multirow{2}{*}{Scene 4} & Trans. (m) & \multicolumn{2}{c}{12.992} & \multicolumn{2}{c}{23.100} & \multicolumn{2}{c}{2.345} & \multicolumn{2}{c}{{\textbf{0.880}}}\\
& Rot. (deg) & \multicolumn{2}{c}{5.445} & \multicolumn{2}{c}{13.088} & \multicolumn{2}{c}{{\textbf{2.872}}} & \multicolumn{2}{c}{3.176}\\ \bottomrule
\end{tabular}
\label{Table_Comparison}
\end{table}
\setlength{\belowcaptionskip}{-20pt}

We also perform ablation experiments to verify the proposed modules of our system. Three versions are used for comparison, namely the complete system (Night-Rider), Night-Rider without the match extension module (Night-Rider w/o ME), and Night-Rider without visual measurements (Night-Rider w/o Vision). Fig. \ref{Ablation} shows the position components of trajectories in four scenes. For Night-Rider w/o Vision, the estimated pose drifts over time, especially in the $z$-axis. With the proposed camera observation model in Night-Rider, the drift is significantly reduced. The trajectories of Night-Rider w/o ME deviate from the true poses, especially in the $z$-axis, since lots of distant streetlight observations are too small to be detected without the match extension module. More 2D-3D streetlight correspondences generated by the match extension module improve the accuracy and robustness of the whole system. The quantitative results in Table. \ref{Table_Comparison} further verify the importance of the camera observation model and the match extension module.

\subsection{Tracking Lost and Degeneration Cases} \label{tracking_lost_and_degeneration_cases}
\begin{figure}[htbp] 
    \centering 
    \includegraphics[width=0.48\textwidth]{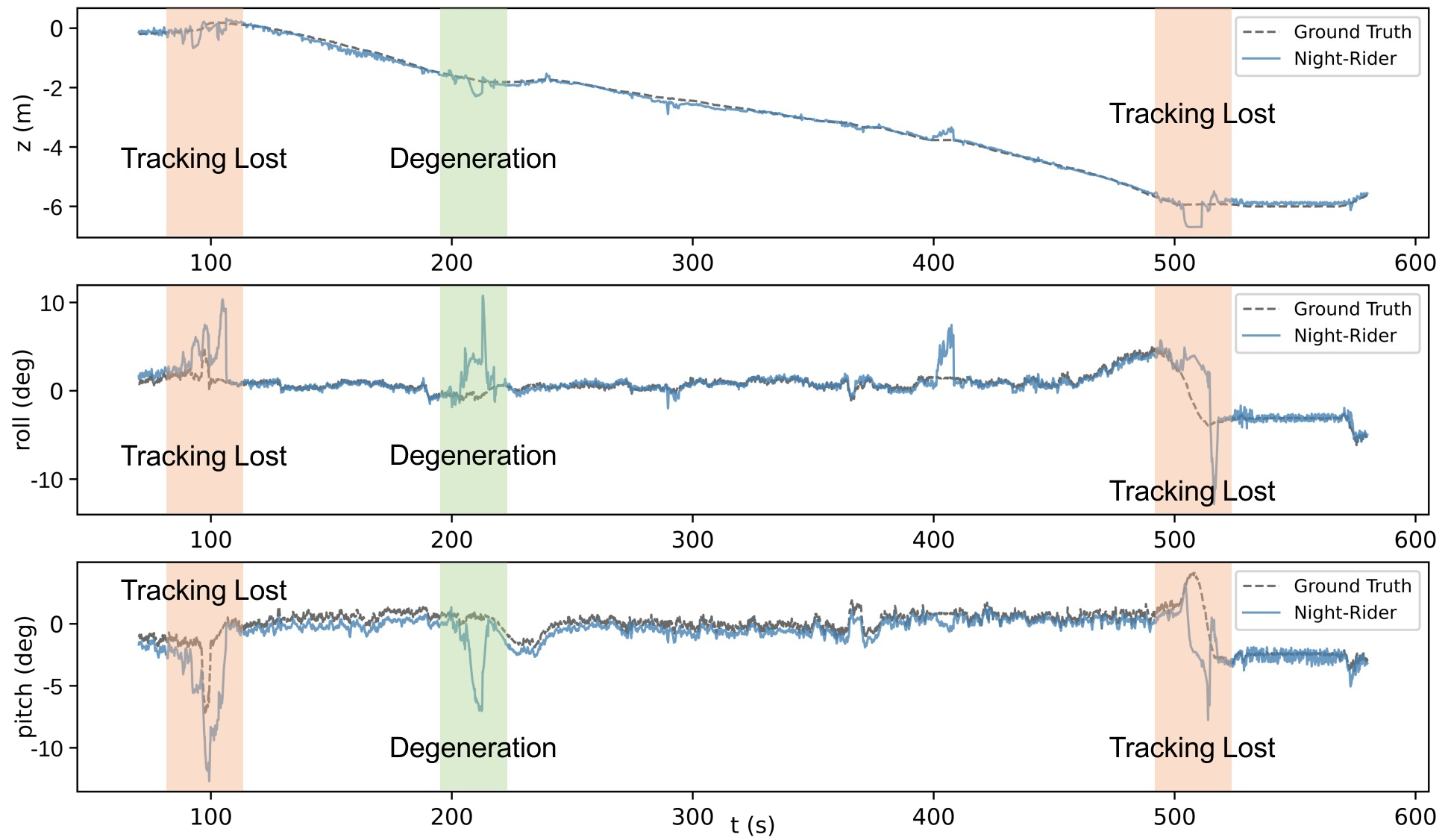} 
    \caption{Cases of tracking failure and degeneration. The proposed system can successfully recover the drifted pose.} 
    \label{Reloc_degen}
    \vspace{0.2cm}
\end{figure}

Fig. \ref{Reloc_degen} shows the estimated trajectory of scene $3$ when the tracking lost or degeneration occurs. Here we display the $z$, roll, and pitch components of poses since the $x$, $y$, and yaw components accurately fit with the ground truth. The system without dealing with tracking lost or degeneration cases fail to estimate the robot poses. It can be observed that when tracking lost occurs (\textcolor{orange}{orange} areas), the robot pose deviates significantly from the true value. Thanks to the proposed tracking recovery module, the pose can be immediately recovered to be close to the ground truth when streetlight observations arrive. The \textcolor{green}{green} area in Fig. \ref{Reloc_degen} indicates that the robot experiences degenerated cases which result in pose drifting over time. When the robot observes streetlights without degeneration cases, the proposed strategy can effectively correct the drifted pose. 

\section{Conclusion}
In this paper, a nocturnal vision-aided localization system integrating camera, IMU, and odometer sensors is proposed for robot navigation at night. To address problems of nocturnal visual localization, we propose to leverage streetlight detections for nocturnal localization. We utilize the InEKF for state estimation and propose novel data association and match extension modules to fully exploit the information from streetlights. Experiments on real nighttime scenes validate that the proposed system achieves accurate and robust localization. Future work includes the integration with mainstream methods and other sensors for a more complete localization system which is adaptively available in both daytime and nighttime scenes. 

\section*{Acknowledgement}
This work was supported in part by the Science and Technology Development Fund of Macao SAR under Grants 0046/2021/AGJ, 0067/2023/AFJ, 0123/2022/AFJ, and 0081/2022/A2.

{
\bibliographystyle{IEEEtran}
\IEEEtriggeratref{14}
\bibliography{references}
}

\end{document}